\documentclass[runningheads]{llncs}
\usepackage{amssymb,amsmath,amsfonts}
\usepackage{algorithmic}
\usepackage{multirow}
\usepackage{multicol}
\usepackage[flushleft]{threeparttable}
\usepackage{makecell}
\usepackage{graphicx}
\usepackage{textcomp}
\usepackage{xcolor}
\usepackage{wrapfig}
\usepackage{algorithm}
\usepackage{algorithmic}
\usepackage{newfloat}
\usepackage{listings}
\usepackage{booktabs}
\floatstyle{ruled}
\newfloat{listing}{tb}{lst}{}
\floatname{listing}{Listing}
\usepackage{fancyvrb}
\usepackage{dsfont}
\usepackage{fvextra}

\usepackage[hidelinks]{hyperref}
\usepackage{amsthm}
\theoremstyle{definition}
\newtheorem{instructions}{Steps}
\newcommand{\Rho}{\mathrm{P}}
\usepackage[misc,geometry]{ifsym}
\begin{document}
\title{ 
Unveiling the Decision-Making Process in Reinforcement Learning with Genetic Programming \\
}
\author{Manuel Eberhardinger \Letter\inst{1}\orcidID{0009-0009-2897-9250}
\and Florian Rupp \inst{2}\orcidID{0000-0001-5250-8613}
\and Johannes Maucher \inst{1}\orcidID{0000-0002-3804-8937}
\and Setareh Maghsudi \inst{3}\orcidID{0000-0002-0647-611X}
}
\authorrunning{M Eberhardinger et al.}
\titlerunning{Unveiling the Decision-Making Process in RL with GP}
%
%
\institute{Hochschule der Medien, Nobelstr. 10, 70569 Stuttgart, Germany 
\email{\{eberhardinger,maucher\}@hdm-stuttgart.de} \\
\and{Hochschule Mannheim, Paul-Wittsack-Straße 10, 68163 Mannheim, Germany 
\email{f.rupp@hs-mannheim.de}} \\
\and{Ruhr-Universität Bochum, Universitätsstraße 150, 44801 Bochum, Germany 
\email{setareh.maghsudi@rub.de}} 
}

\maketitle    
\begin{abstract}
Despite tremendous progress, machine learning and deep learning still suffer from incomprehensible predictions. Incomprehensibility, however, is not an option for the use of (deep) reinforcement learning in the real world, as unpredictable actions can seriously harm the involved individuals. 

In this work, we propose a genetic programming framework to generate explanations for the decision-making process of already trained agents by imitating them with programs. Programs are interpretable and can be executed to generate explanations of why the agent chooses a particular action. Furthermore, we conduct an ablation study that investigates how extending the domain-specific language by using library learning alters the performance of the method. We compare our results with the previous state of the art for this problem and show that we are comparable in performance but require much less hardware resources and computation time. 
\end{abstract}
\keywords{Genetic Programming \and Explainable Reinforcement Learning}

\section{Introduction}
In recent years, machine learning and deep learning have made enormous progress both in research and in everyday use to support users in all kinds of tasks. While they are helpful in the short term, there is still no guarantee that the predictions or the generated content will be correct.
Explainable artificial intelligence (XAI) is on the rise to alleviate some of the problems in understanding how a model generates a prediction. In the case of decision-making and reinforcement learning (RL), it is even more important to understand how an agent makes a decision. Currently, RL is hardly applicable in the real world \cite{dulac-arnold_challenges_2021}, since unpredictable actions can cause accidents with potentially serious and long-lasting consequences for the involved individuals.

Creating comprehensible explanations for the decision-making process of agents is still an underexplored problem. 
Recent work \cite{eberhardinger_learning_2023} showed that it is possible to generate post-hoc explanations for the decision-making process with program synthesis in combination with library learning. The objective of program synthesis is to find a program for a given specification, such as input-output examples or a natural language description \cite{gulwani_program_2017}. Library learning benefits program synthesis by extracting functions from already synthesized programs, thus, generating short-cuts in the program search space. This leads to faster search and finding of more programs in the same amount of time \cite{ellis_dreamcoder_2021}. In addition, using programs to control an agent has the benefit that the program can be verified before using it in a production system \cite{lathouwers_modelling_2022} or can be adapted to other scenarios by software engineers \cite{trivedi_learning_2022}.

\begin{figure}[tb]
  \begin{center}
  \includegraphics[width=\linewidth]{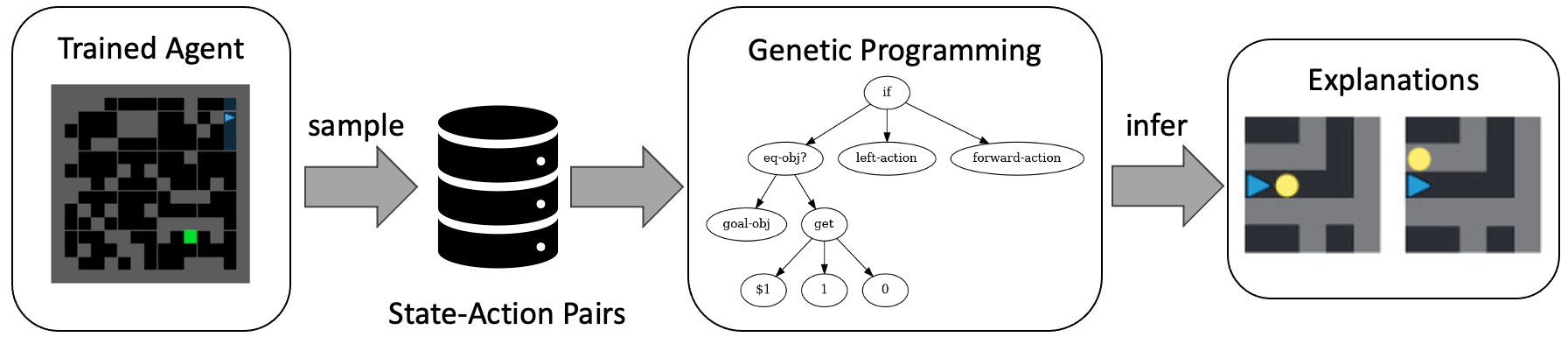}
    \end{center}
  \caption{The overview of the problem setting. We train an agent to sample state-action pairs from the learned policy. These examples are then imitated with the genetic programming algorithm to generate explanations for the agent's decision-making process.}
    \label{fig:overview}
\end{figure}

This work studies how genetic programming (GP) \cite{koza_genetic_1992} can alleviate some of the drawbacks of the proposed method from \cite{eberhardinger_learning_2023}, by first conducting a study, what the problems of the different kind of program synthesizers for these tasks are and how GP tackles those problems. Afterwards, we introduce a GP algorithm which can generate explanations for grid-based RL environments, and show the feasibility of the method for explaining decisions of a maze running agent. This algorithm combines a typed domain-specific language based on Lisp \cite{mccarthy_recursive_1960} in combination with library learning. 
Figure \ref{fig:overview} shows an overview of the problem setting and approach. We train an agent to sample sub-trajectories, i.e. state-action pairs, from the learned policy. These examples are then imitated with the GP algorithm to generate explanations for the agent's decision-making process. This is possible by highlighting the positions on the state, which the agent checks in the program.

Our contributions include:
\begin{itemize}
    \item A genetic programming algorithm for explaining decisions of reinforcement learning agents.
    \item Ablation studies on library learning for genetic programming.
    \item An experimental study to evaluate the proposed genetic programming algorithm to explain decisions of an agent solving a maze.
    \item The code is open source and available on Github\footnote{\url{https://github.com/ManuelEberhardinger/unveiling-rl-with-gp}} for further research.
\end{itemize}

The structure of the paper is as follows: Section \ref{ref:related-work} gives a brief overview of related work in the area of genetic programming and explainable RL. Afterwards (\ref{ref:background}) the background of this work is presented with the case study of the drawbacks of the previous proposed method. In Section \ref{ref:method} the method is described, followed by the experiments (\ref{ref:experiments}). The discussion is in Section \ref{ref:discussion}, the conclusion in \ref{ref:conclusion}.

\section{Related work}
\label{ref:related-work}
Genetic programming \cite{koza_genetic_1992} and especially program synthesis \cite{waldinger_prow_1969} have a long history in the computer science research community. 
The application of GP to improve the interpretability of various machine learning models has been studied before in~\cite{ferreira_applying_2020}. Therefore, the authors introduce GPX, the Generic Programming Explainer, which generates a tree structure of the behavior of models like deep neural networks, random forests or support vector machines.
Another approach to explain a learned RL policy was introduced in~\cite{videau_multi-objective_2022}, which is based on a multi-objective GP algorithm. The authors achieve state of the art results when comparing the interpretable extracted policy in different control environments. In~\cite{liventsev_neurogenetic_2021} a neurogenetic approach is implemented to directly learn interpretable reinforcement learning policies. The method was applied to common OpenAI Gym environments and was able to keep up with the leaderboards.
Another approach, genetic programming for reinforcement learning (GPRL), is introduced by Hein et al.~\cite{hein_interpretable_2018}. The method is able to extract compact algebraic equations or boolean logic terms from observed trajectories.



However, GP is not the only method in the field of explainable reinforcement learning (XRL). In a recent extensive survey \cite{milani_explainable_2023}, XRL is categorized into three explainable facets: feature importance, learning process and Markov decision process, and policy-level.
Among these categories, feature importance provides explanations for which features of the state affected the decision-making process of the agent. This category includes programmatic policies, where a policy is represented by a program, since programs directly use the state of the environment and thus the program can be traversed to generate an explanation \cite{verma_imitationprojected_2019,verma_programmatically_2018,anderson_neurosymbolic_2020,qiu_programmatic_2022}.
Furthermore, other works within this category synthesize finite state machines to represent policies \cite{inala_synthesizing_2020} or use decision tree-based models \cite{silver_fewshot_2020,bastani_verifiable_2018}. Our methodology fits into this category as we explain sub-trajectories of policies and our overarching goal in the future is to extract a program capable of representing the complete policy.

\section{Background}
\label{ref:background}
In this section, we introduce the various definitions and ideas on which our work is based. We first provide a general overview of the domain-specific language and then give a detailed description of the framework proposed in \cite{eberhardinger_learning_2023}.

\subsection{Program and Domain-specific Language}
\begin{table}[t]
    \caption{The used domain-specific language at the beginning. The type column shows one type for values and several types separated by an arrow for functions. The type after the last arrow is the return type of the function. The types before it are the types of the input parameters. }
    \centering
    \resizebox{\linewidth}{!}{
    \begin{tabular}{llll}
         Functions/Values & Description & Type  \\ \hline
         left, right, forward & possible actions & action   \\ 
         0, 1, ...,  5 & integer values & int \\
         \$0  & the direction the agent is facing & agentDirection \\
         \$1  & 2D grid observation of the agent & map \\
         direction-\{0, ..., 3\} & represents north, east, south and west & direction \\ 
         wall, empty, goal & possible objects on the map & object \\
         mapObject  & a object on a map with coordinates & mapObject \\
         if & standard if-clause & bool $\small\rightarrow$ func $\small\rightarrow$ func $\small\rightarrow$ func\\
         eq-direction? & checks if two directions are equal & agentDirection $\small\rightarrow$ direction $\small\rightarrow$ bool \\
         eq-obj? &  checks if two objects are equal & mapObject $\small\rightarrow$ object $\small\rightarrow$ bool \\
         get & get a object on the map & map $\small\rightarrow$ int $\small\rightarrow$ int $\small\rightarrow$ mapObject \\
         get-game-obj & get the object type of a mapObject & mapObject $\small\rightarrow$ object  \\
         not & negates a Boolean value & bool $\small\rightarrow$ bool \\
         and &  conjunction of two Boolean values & bool $\small\rightarrow$ bool $\small\rightarrow$ bool \\
         or &  disjunction of two Boolean values & bool $\small\rightarrow$ bool $\small\rightarrow$ bool \\
    \end{tabular}}
    \label{table:dsl}
\end{table}
To represent programs, this work uses a typed domain-specific language (DSL), inspired by the Lisp programming language \cite{mccarthy_recursive_1960}, represented as a uniformly distributed probabilistic grammar, i.e., each production rule has the same probability \cite{manning_foundations_1999}. The DSL's core components consist of control flow production rules, the available actions the agents can use and modules designed to help the agent perceive its environment. Given our focus on grid environments, the agent's perception relies on modules that determine specific positions on the grid and compare them with possible objects of the environment such as walls, empty cells, or game-specific objects. Additionally, the control flow production rules consists of if-else statements and Boolean operators, enabling the synthesis of more complex conditions. The full DSL is shown in Table \ref{table:dsl}. 

As we use a typed DSL, we explicitly denote types for every function or constant. Constants are indicated with a single type in the type column. For functions, multiple types are linked using an arrow symbol $\small\rightarrow$. The last type is the return type of the function, the preceding types are the input parameters. To minimize the runtime of the type checks, we use a specific if-production-rule for each possible type in the implementation, i.e. we have a specific if-clause that only returns actions of the agent, another that only returns game objects and so on.

\begin{listing}[h]%
\caption{An example program generated from the domain-specific language.}%
\label{lst:listing}%
\begin{lstlisting}[language=Java, escapeinside={(*}{*)}]
(*$\lambda$*)(x) ( 
  (if (eq-obj? goal-obj (get x 1 0))
        left-action forward-action)
)
\end{lstlisting}
\end{listing}

Listing \ref{lst:listing} shows an example program generated from the DSL that has an input parameter x representing a grid environment and checks whether the position on (1,0) is the goal position. If this is the case, the left action is chosen, otherwise the forward action is chosen.

\subsection{Program Synthesis with Library Learning for Reinforcement Learning}\label{sec:aiide}
In \cite{eberhardinger_learning_2023} a method is introduced to learn generalizable and interpretable knowledge in grid-based RL environments. To achieve this, the authors adapt the state-of-the-art program synthesis system DreamCoder \cite{ellis_dreamcoder_2021} for reinforcement learning. First, RL agents are trained for a given environment to collect data to imitate. This is equivalent to sampling of the state-action pairs from the learned policy in Figure \ref{fig:overview}. After the data has been collected from the trained agents, the program synthesis part begins. 
This part consists of an iterative procedure which is guided by a curriculum \cite{bengio_curriculum_2009} that provides the program synthesizer with the input-output examples. The curriculum is able to adapt the sequence length of the sub-trajectories to make the task more challenging and is based on the intuition that shorter sequences are easier to imitate than longer ones. Shorter sequences contain less information, so the programs are shorter and easier to synthesize. 

The iterative procedure consists of three possible components: the search for programs for the provided state-action pairs from the curriculum which can be done with a neural network or a symbolic approach; the generation of a training dataset and the training of the neural network; a library learning module that analyzes programs that can imitate at least one sub-trajectory to extend the DSL. 
In \cite{eberhardinger_learning_2023} three different kinds of synthesizers are studied, a language model (LM) fine-tuned on code \cite{wang_codet5_2021}, a brute-force enumerative search and a neural-guided enumerative search \cite{ellis_dreamcoder_2021}. 
These program synthesizers are integrated into the iterative procedure that is guided by the curriculum. The necessary steps for each method are described briefly below.

\begin{instructions}\label{steps:dreamcoder}
{For the enumerative search and the neural-guided search theses steps are necessary:}
\begin{enumerate}
    \item Perform a brute-force search that enumerates programs in decreasing order of the probability of being generated from the probabilistic grammar until a timeout is reached or a program is found. The enumerated programs are  checked if they can solve any of the provided state-action pairs.
    \item Generate random training data in addition to the found programs from step 1 and train a neural network which predicts probabilities for all production rules in the grammar. This leads to programs being found faster as they are checked earlier in the enumerative search. This step is only used for the neural-guided search. 
    \item The found programs are sent to the library learning module, where they are refactored and checked for recurring semantic patterns, to extract functions from \cite{ellis_dreamcoder_2021}.
    \item Update the input-output examples with the provided state-action pairs from the curriculum. Then start again at the first step with the brute-force search. 
\end{enumerate}
\end{instructions}

\begin{instructions}\label{steps:lm}{For the language model similar steps are necessary:}
\begin{enumerate}
    \item Generate a random training dataset of 50000 state-action pairs by sampling random programs and executing them on a given environment.
    \item Train CodeT5 \cite{wang_codet5_2021} on the generated dataset.
    \item Synthesize programs for the test data collected from an agent. 
    \item Send correct programs to the library learning module to extend the DSL.
    \item Update the input-output examples with the provided state-action pairs from the curriculum. Then start again at the first step with generating a random training dataset since the DSL has been updated.
\end{enumerate}
\end{instructions}

\subsection{Case Study: Drawbacks of previous Method}
\label{case-study}
All proposed program synthesizers of \cite{eberhardinger_learning_2023} have several disadvantages compared to a GP algorithm for this task. Both enumerative search methods always start from scratch of the search process instead of exploiting the fact that the input-output examples are provided by the curriculum and thus share similar structures. GP takes advantage of this by making local changes and keeping the population of the last iteration as the starting population for the next iteration.

The greatest issue with an LM is the dataset generation. In \cite{eberhardinger_learning_2023}, 50000 random programs are generated in each iteration and are executed in the RL environment. However, recent work has shown that 50000 random programs are far from sufficient to learn the semantics of the DSL \cite{jin_evidence_2023}. Jin et al. showed that for a simplified DSL, without if-clauses and Boolean operators, the LM learns the semantics of a DSL, which controls an agent in a grid environment \cite{jin_evidence_2023}. To achieve this, the authors generated one million programs for the training dataset, 20 times the size of the dataset to train CodeT5. 

In addition, the library learning module of DreamCoder has a high memory and time requirement. These drawbacks have been mitigated by a new library learning algorithm called Stitch  \cite{bowers_top-down_2023}, which is about three orders of magnitude faster and requires about two orders of magnitude less memory. DreamCoder uses a deductive approach that is challenging to scale to longer programs because programs are refactored with rewrite rules, resulting in huge memory and time requirements. Stitch, on the other hand, synthesizes abstractions directly through a method called corpus-guided top-down synthesis, which uses a top-down search to constrain the search space and guide it to abstractions that cover most of the common syntactic structures of all provided programs, without the need to refactor the programs \cite{bowers_top-down_2023}.
In this work, we use the Stitch module instead of the DreamCoder library learning module.

\section{Methodology}
\label{ref:method}
In this section, we present a genetic programming algorithm for explaining the decision-making process of RL agents. First, we give an overview of the general algorithm, including the mutation and crossover operators. Second, the integrated curriculum, bloat control and the library learning module are described. 

\subsection{Genetic Programming}
\begin{figure}[tb]
  \begin{center}
  \includegraphics[width=0.6\linewidth]{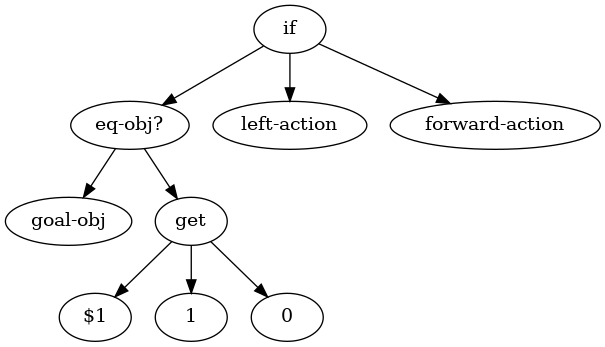}
    \end{center}
  \caption{The abstract syntax tree for the example program from Listing \ref{lst:listing}.}
    \label{fig:example-tree}
\end{figure}
Given that our DSL is based on Lisp, the proposed GP algorithm is tree-based and directly modifies the abstract syntax tree of Lisp \cite{koza_genetic_1992}. Figure \ref{fig:example-tree} shows the corresponding tree for Listing \ref{lst:listing}. The various operators and the fitness function for the genetic algorithm are described below: 

\textbf{Initialization of Population:}
We initialize the population $\Rho$ with random programs by sampling them from the uniformly distributed grammar. Since we use a typed DSL, the programs are sampled by specifying the return type of the program. We randomly select a production rule for the given return type and then sample subtrees for the input parameters of that production rule. This is repeated until all leaves of the tree are terminals, i.e., have no input parameters.

\textbf{Mutation:}
With a probability of $p_{mutation}$ per node, we randomly choose nodes for mutation. We then select a random production rule for the same return type as the selected node and check whether the input parameter types match. If they match, we keep the child nodes of the tree, if not, we sample new subtrees for each child parameter. This mutation process is similar to sampling random programs for initializing the population, as only the return type is specified to generate a new tree. 

\textbf{Crossover:}
We use a one-point crossover operator where we randomly select a node from a given tree of a program with a probability of $p_{crossover}$ for each node. We then scan the other tree for the same return type and randomly select one of those nodes. Afterwards, we link the subtree to the parent of the selected node from the other tree and do it for the remaining node vice versa.

\textbf{Fitness:}
Fitness is evaluated based on the number of correct state-action pairs that the algorithm imitates with an individual. We adapt the formula for calculating the accuracy for a set of programs from \cite{eberhardinger_learning_2023} to calculate the proportion of correctly solved tasks for a single program $\rho$, i.e. an individual in the population. Equation \ref{eq:fitness} calculates the fitness of an individual for the dataset $D$ to imitate. $N_D$ is the size of $D$, i.e. the number of state-action pairs. A sub-trajectory from $D$ is defined by $\tau$, which consists of state-action pairs $(s,a)$.
Equation \ref{eq:exec} defines $f(\rho,\tau)$ that evaluates a program $\rho$ and
returns 1 if the complete rollout $\tau$ can be imitated and otherwise 0.
$\operatorname{EXEC}(\rho, s)$ executes the program for the state $s$  and returns an action $a$.
Boolean values are mapped to 0 and 1 with the identity function $\mathds{1}$.

\begin{equation}
\label{eq:fitness}
Fitness = \frac{1}{1+ N_D - \sum_{\tau \in D} f(\rho, \tau)},
\end{equation}
\begin{equation}
\label{eq:exec}
f(\rho,\tau) = { \underbrace{ \mathds{1} \left\{ \text{ }\operatorname{EXEC}(\rho,s)==a,  \text{         } \forall (s,a) \in \tau \text{     }\right\},}_{\text{is } 0 \text{ after the first } (\rho,s) \text{ where } \operatorname{EXEC}(\rho, s) \text{ } != \text{ }  a}} 
\end{equation}

\textbf{Selection:} 
Tournament Selection \cite{miller_genetic_1995} is used to select individuals for mutation and crossover. First, $k$ individuals are selected randomly and then from this smaller population, the individual with the best fitness is selected.

\textbf{Curriculum:}
The curriculum is integrated into the GP algorithm and increases the sequence length of the state-action pairs every 10 generations or if at least 95\% of the collected data is imitated. Afterwards, the library learning module is used to analyze all correct programs.
The curriculum stops when no individual in the population is able to imitate a state-action pair. 

\textbf{Bloat Control:}
Bloat control, which limits the excessive growth of the program trees in the population, and simplification of programs is an important aspect in GP to make programs interpretable and concise \cite{javed_simplification_2022}. For this reason, we limit the growth of the program trees by integrating a bloat control into the fitness function. For this, we need an auxiliary function $size$, which returns the size of the input program, i.e. how many production rules the program has used. The following equation calculates the final fitness for an individual $\rho$:
\begin{equation}
    Fitness = \frac{1}{1+N_D - \sum_{\tau \in D} f(\rho, \tau) + w_b * size(\rho)},
\end{equation}
where $w_b$ is the bloat weight, which penalizes bigger programs. 

It is also possible to enforce the simplification of programs with a more restrictive type system in the DSL, resulting in more concise programs without excessive bloat. The types of the functions can be more restricted, for example by only allowing the comparison of the direction of the agent \verb|agentDirection| with the normal direction type \verb|direction| in the method \verb|eq-direction?|, which can be seen in the type column of the Table \ref{table:dsl}. The same type restrictions are also taken into account when comparing possible objects on the map. This removes unnecessary Boolean conditions from the sampling process of random programs or in mutations, resulting in shorter and more concise programs. 

\subsection{Library Learning}
In this step, the library learning module analyzes all programs, which could solve at least one task, and extract functions from these programs to extend the DSL. Before we add a new function to the DSL, we check whether the size of the function is below a predefined value. In our experiments, we used 10 to mitigate the problem that abstractions are too specific and the search gets stuck in local minima. This is similar to pruning of decision trees \cite{esposito_comparative_1997}. In addition, the number of extracted functions is limited, as otherwise too many production rules in the DSL could hinder the search \cite{cropper_forgetting_2020}. To enhance the execution speed of the extracted functions, we implement a cache that stores the return value for each input parameter configuration.

\section{Experiments}
\label{ref:experiments}
In this section, we evaluate the proposed algorithm. First, the domain is explained, then the execution time and accuracy are compared with the baseline methods. Afterwards, we conduct an ablation study on the influence of library learning for genetic programming.

\subsection{Domain}
\begin{wrapfigure}{r}{0.25\linewidth}\vspace{-30pt}
  \includegraphics[width=\linewidth]{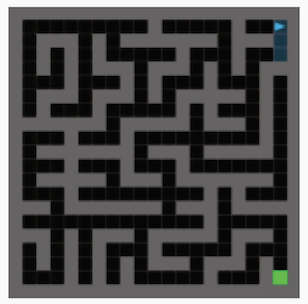}
  \caption{The medium sized perfect maze environment which is used to evaluate the proposed method.}
    \label{fig:minigrid}\vspace{-30pt}
\end{wrapfigure}
In this work, we use the grid-world domain \cite{chevalier-boisvert_maxime_minimalistic_2018} and a maze-running task with a partial observation of 5x5 from the agent's point of view. We trained an agent to collect data with the default hyperparameters provided by \cite{parker-holder_evolving_2022}. Imitation learning is necessary due to the full size of the maze, the partial observation of the agent and the sparse rewards that are only given if the agent finds the exit of the maze. This makes it challenging to solve this problem directly with GP. 
Figure \ref{fig:minigrid} shows the medium-sized perfect maze environment used to collect state-action pairs for the imitation learning task. The agent's task is to find the exit of the maze. The available actions are moving forward, turn left or turn right.

\subsection{Evaluation}
We compare the runtimes and accuracy of the proposed method to the program synthesizers from \cite{eberhardinger_learning_2023}: enumerative search, DreamCoder, CodeT5 and LibT5, CodeT5 with a library learning module.

In our experiments, we use the same hyperparameters for all genetic programming runs, except for the hyperparameter whether library learning is used. The population size is 1000, the tournament size is 100,  $p_{mutation}=0.5$, $p_{crossover}=0.5$ and $w_b=0.025$. We also limit the maximum depth of sampling random programs from the grammar to six. The number of generations is limited by the curriculum and is therefore not explicitly specified.

\subsubsection{Runtime Improvements}
\begin{table}[t!]
    \centering
    \caption{The runtimes of the evaluated methods for generating explanations of the decision-making process of a maze 
 runner agent. Phase 1 refers to the training phase and Phase 2 refers to the time for synthesizing programs for the test tasks. A more detailed explanation of the individual phases is given in the main text. All reported numbers for Phase 1, Phase 2, Library Learning and Sum are in minutes. }
    \begin{threeparttable}
    \begin{tabular*}{\linewidth}{@{\extracolsep{\fill}}clrrr!{     \thinspace  \vrule  }r}
    \toprule
    Sequence Length     &  Method           & Phase 1 & Phase 2 &  Library Learning &  Sum \\
    \midrule
    \multirow{3}{*}{3}  & GP                & -     &  0.82     & 0.59              & \textbf{1.41}  \\
                        & DreamCoder        &  9.19 &  11.23    & 54.07             & 74.49 \\
                        & Language Model    & 56.06 &  12.71    &  4.84             & 73.60 \\
    \midrule
    \multirow{3}{*}{4}  & GP                & -     &  1.07     &  0.63             & \textbf{1.70} \\
                        & DreamCoder        & 1.73  &  5.13     &  1.37             & 8.23 \\
                        & Language Model    & 57.45 &  13.31    & 2.28              & 73.04 \\
    \midrule
    \multirow{3}{*}{5}  & GP                & -     &  1.44     &  0.64             & \textbf{2.08} \\
                        & DreamCoder        & 12.00 &  14.41    & 12.60             & 39.02 \\
                        & Language Model    & 59.93 &  11.92    &  13.54            & 85.39 \\
    \midrule
    \multirow{3}{*}{6}  & GP                & -     &  1.39     &  0.68             & \textbf{2.08} \\
                        & DreamCoder        & 12.00 &  14.93    &  17.88            & 44.81 \\
                        & Language Model    & 61.88 &  10.58    &  5.81             & 78.28 \\
    \midrule
    \multirow{3}{*}{7}  & GP                & -     &  1.18     &       0.69        & \textbf{1.87} \\
                        & DreamCoder        & 12.00 &  13.78    &  20.35            & 46.14 \\
                        & Language Model    & 62.80 &  9.05     &  1.25             & 73.10 \\
    \midrule
    \multirow{3}{*}{8}  & GP                & -     &  1.36     &              0.72 & \textbf{2.08} \\    
                        & DreamCoder        & 12.00 & 13.83     &              4.65 & 30.49 \\
                        & Language Model    & 64.31 &  8.13     &              2.02 & 74.47 \\
    \midrule
    \multirow{3}{*}{9}  & GP                & -     &  1.68     &              0.76 & \textbf{2.45} \\
                        & DreamCoder        & 12.00 &  13.78    &             26.10 & 51.89 \\
                        & Language Model    & 64.87 & 7.66      &              3.11 & 75.64 \\
    \midrule\midrule
    \multirow{5}{*}{\makecell{Cumulative Sum of \\ all sequence lengths}}  & \multicolumn{4}{l}{GP}      & \textbf{13.67} \\
                        & \multicolumn{4}{l}{Enumerative Search\tnote{1}}  &  208.00 \\
                        & \multicolumn{4}{l}{DreamCoder} & 295.08 \\
                        & \multicolumn{4}{l}{CodeT5\tnote{2}}   & 500.66 \\
                        & \multicolumn{4}{l}{LibT5\tnote{3}}  & 533.52 \\
                        
    \bottomrule
    \end{tabular*}
      \begin{tablenotes}
        \item[1] Phase 1 and Library Learning of the DreamCoder runtimes.
        \item[2] Phase 1 + Phase 2 of the Lanuage Model runtimes.
        \item[3] All phases of the Language Model runtimes.
      \end{tablenotes}
    \label{tab:runtime}
    \end{threeparttable}
\end{table}
To show the runtime improvements of GP in the training phase, the time is measured until the threshold for increasing the sequence length is reached. We start by imitating state-action pairs with a sequence length of three and report the runtimes for each iteration until a sequence length of ten is reached, as the times can be extrapolated to the larger sequence lengths and different sequence lengths are reached at the end depending on the method. To ensure the fairest possible comparison, each method uses the same state-action sequences to imitate and similar CPU resources. For GP, enumerative search and DreamCoder, we use four CPU cores from the AMD EPYC processor family 23 model 1. For the language models, four CPU cores from the  Intel(R) Core(TM) i7-6950X CPU @ 3.00GHz family 6 model 79 in combination with a 48GB NVIDIA GeForce RTX 2080 Ti GPU are used.

For this experiment the number of programs for DreamCoder's library learning module is restricted to 50, as otherwise it will run out of RAM. 
We exclude CodeT5 and the enumerative search from the table, except for the cumulative sum of all sequence lengths, since the enumerative search is also part of DreamCoder and CodeT5 has the same runtime as LibT5, but without the library learning module. To increase the sequence length, we use the identical curriculum strategy as described in \cite{eberhardinger_learning_2023}. This strategy extends the sequence length when at least 10\% of the state-action pairs are imitated and halts if no sequences can be imitated, for two consecutive instances.

Table \ref{tab:runtime} compares the different program synthesizers based on their runtimes. For GP there exists no phase 1 since we do not need to train a neural network; phase 2 is the runtime of the GP algorithm introduced in section \ref{ref:method}. Phase 1 for DreamCoder refers to steps \ref{steps:dreamcoder} point 1 and phase 2 to steps \ref{steps:dreamcoder} point 2 in section \ref{sec:aiide}. 
Phase 1 of the language model are the steps \ref{steps:lm} point 1 and 2; phase 2 refers to steps \ref{steps:lm} point 3 in section \ref{sec:aiide}.
The proposed method outperforms all other methods by a large margin. We hypothesize that the runtime drop in DreamCoder for sequence length four is due to the fact that the tasks are too simple after extracting a library in the previous iteration, resulting in fast search times since the search stops when all tasks are imitated.

In the last row of Table \ref{tab:runtime} the corresponding cumulative sum of the runtimes for each evaluated method is shown in minutes. GP outperforms all other methods by more than an order of magnitude.

\subsubsection{Accuracy}
\begin{figure}[b]
  \begin{center}
  \includegraphics[width=0.7\linewidth]{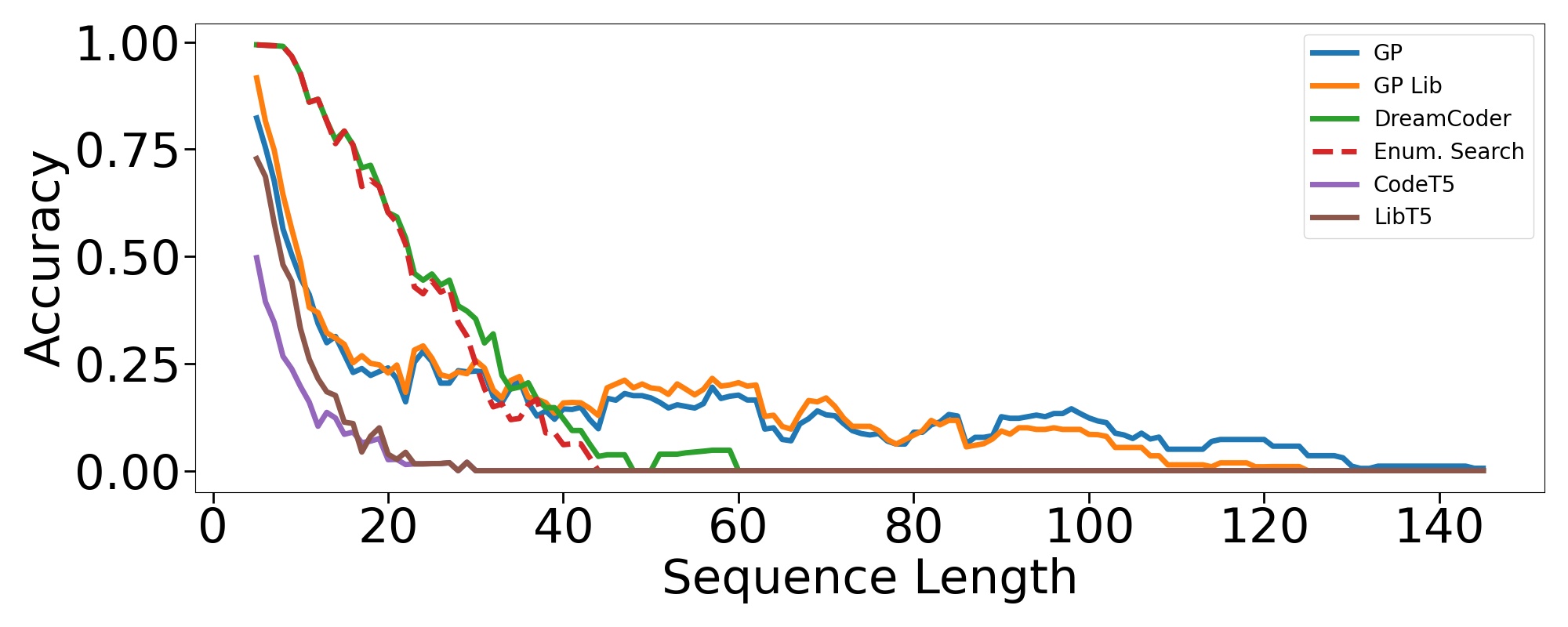}
    \end{center}
  \caption{The evaluation of the genetic programming algorithm compared to the baseline methods. The accuracy is displayed on the y-axis and the the sequence length on the x-axis.}
    \label{fig:eval-acc}
\end{figure}
Figure \ref{fig:eval-acc} shows the accuracy of the various program synthesizers. The accuracy calculates the proportion of correctly imitated state-action pairs available for each sequence length \cite{eberhardinger_learning_2023}. 
For the genetic algorithm, the results are averaged over ten runs using different seeds.
For shorter sequences up to 35, GP is in the middle between DreamCoder and the neural program synthesizers, but offers runtime improvements and lower hardware requirements.
GP is the best method for longer sequences, as it can imitate sequences twice as long as DreamCoder. 
The symbolic-based approaches have a big margin over the LM-based program synthesizers.


\subsubsection{Library Learning}
\begin{figure}[tb]
  \begin{center}
  \includegraphics[width=0.7\linewidth]{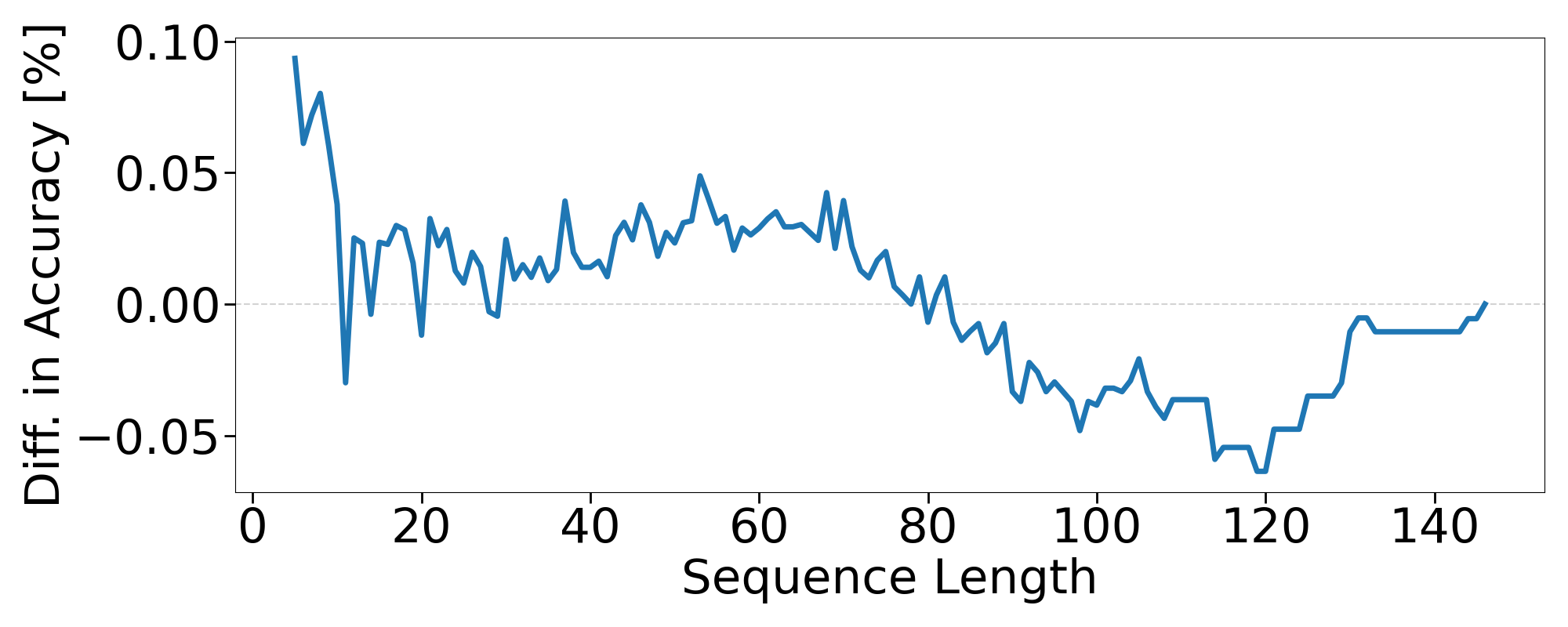}
    \end{center}
  \caption{The difference between using library learning with genetic programming and without.}
    \label{fig:eval-lib}
\end{figure}
To investigate the influence of learning a library of reusable functions in combination with GP, we plot the difference in accuracy for a genetic programming run with and without library learning in Figure \ref{fig:eval-lib}. For shorter sequences, almost 10\% better accuracy is achieved. For sequence lengths from 10 to 80, library learning is better in most cases. For sequences longer than 80, library learning is detrimental. 

\begin{listing}[tb]%
\caption{The first five extracted functions of Stitch. For the sake of simplicity, we only show the body of the function. All numbers preceded by \# represent input parameters of the function.}%
\label{lst:library}%
\begin{lstlisting}[language=Java, escapeinside={(*}{*)}]
fn_0=(eq-obj? (get #2 #1 #0) #3)
fn_1=(if_action #0 left-action forward-action)
fn_2=(fn_0 #2 #1 #0 (get-game-obj (get #0 2 1)))
fn_3=(eq-obj? (get #0 2 1) wall-obj)
fn_4=(get #0 2 1)
\end{lstlisting}
\end{listing}
In addition, we analyze the functions that have been added to the DSL. Listing \ref{lst:library} shows the first five extracted functions of Stitch. All numbers preceded by \# represent input parameters of the function. It can be seen that high-level concepts are extracted from the synthesized programs. Depending on the function, more or less production rules are saved when using the library instead of the original DSL. The function \verb|fn_3| checks the position (2,1), which is directly in front of the agent, and compares it with a wall object. Using this function in combination with \verb|fn_1| allows the program synthesizer to synthesize a valid program with three production rules, \verb|fn_1|, \verb|fn_3| and the grid observation.
This program checks whether there is a wall in front of the agent and turns left if this is the case, otherwise the agent moves one step forward.

\section{Discussion \& Limitations}
\label{ref:discussion}
In our experiments, we have shown the benefits of using genetic programming for the task of imitating sub-trajectories for an agent solving a grid-based maze environment. There are, however, various aspects we need to discuss for the different experiments. 

When evaluating the program synthesizers, it is difficult to make a fair comparison as they are all based on different methods. 
The enumerative and neural-guided search have a timeout limit of 720 seconds and thus stop after the given amount of time. The neural program synthesizers are allowed to generate 100 programs for each testing task. GP has for every sequence length a maximum of 10 generations. Table \ref{tab:runtime} shows that the specified computation budget for LMs and DreamCoder is comparable. For our GP approach, we need much less computation time and at the same time improve the results for longer sequence lengths than all other methods. Even when considering the worst-case scenario and all 10 generations are required, GP takes 136.7 minutes (19.53 min per generation on average) to enhance the sequence length, yet it remains more than 70 minutes faster than the fastest among the alternative methods. This is one of the main advantages of using GP for this type of task and shows a significant reduction in hardware requirements, resulting in more sustainable computing compared to the other methods. Unlike LMs, no GPU is necessary, which significantly reduces the environmental impact \cite{DEVRIES20232191}. Even though it is possible to achieve human-competitive programming with LMs \cite{li_competition-level_2022}, this is limited to large amounts of available training data, which is not possible in our case since the kind of programs we want to synthesize in a customized DSL are not available anywhere online. 
Other related problems of LM-based synthesizers were already discussed in section \ref{case-study}.

The GP-based methods are not as good as DreamCoder and the enumerative search for the shorter sequences up to a length of 35. 
In terms of the main goal of this research, a policy extraction algorithm, this is not a problem.
It is even more important to imitate longer state-action sequences, since programs for longer sequences are more reliable as they explain more of the policy.
For shorter programs, it is much easier to find a program with a false explanation, i.e. focusing the attention to incorrect grid cells that were not responsible for the decision-making process of the agent.  
With this in mind, we successfully mitigated the drawbacks of DreamCoder with GP described in section \ref{case-study} by reusing the population from the previous iteration, resulting in significantly longer imitated sequence lengths. 
Nevertheless, more work is necessary for a complete policy extraction algorithm, but we think GP is a big step forward compared to the other program synthesizers to enable rapid experimentation for further research.

We were able to improve the integration of library learning in GP by limiting the number of extracted functions to five or adding only functions that are concise, i.e. functions that use less than ten production rules. We hypothesize that adding too many functions steers the search into local maxima by allowing shortcuts in the search space that are hard to escape. Adding too specific functions leads to the same problem. 
The drop in the difference in accuracy for sequence lengths above 80 (see Figure \ref{fig:eval-lib}) is another indication that library learning steers the search into local maxima and the more functions are extracted, the earlier this occurs. If we do not limit the number of functions at all, GP with library learning does not reach a sequence length of 45. 
Another possible reason for this behavior is that functions added in the first iterations for short sequences are not useful for longer sequences. This is the case if the synthesized programs provide incorrect explanations. Consequently, the question arises how the assessment of whether the explanations are correct, can be conducted. A recent study discussed these and other problems related to XAI in general \cite{ali_explainable_2023}. Future work is imperative to assess the trustworthiness of explanations for shorter sequences. 


\subsection{Complexity of Implementation}
Another advantage of this work is the complexity of the implementation, which reduces the effort required for follow-up work and for other researchers who want to use this method for other problems. Our implementation only requires Python, whereas DreamCoder uses OCaml and Python in combination. DreamCoder requires 45050 lines of code for the Python files and 12099 lines of code for the OCaml files. In contrast, the method proposed in this paper requires only 1247 lines of code in total - only 2.18\% of the DreamCoder implementation, so this implementation is much easier to understand and to customize for follow-up research.

\section{Conclusion and Future Work}
\label{ref:conclusion}
In this paper, we have presented a tree-based genetic programming method to explain the decision-making process of a reinforcement learning agent. 
To achieve this, a typed domain-specific language is used that allows to directly modify the abstract syntax tree in the mutation and crossover operators.  
In addition, we have integrated learning a library of functions, that represent high-level concepts the agent has learned, into the framework. 
We have shown improvements in runtime and achieved better accuracy at longer sequence lengths than the baseline methods. 
By using fewer hardware resources than the previous methods, we have also emphasized the sustainable computing aspects of genetic programming compared to neural program synthesis. 

A promising approach for future work is to reduce the complexity of the search problem by considering not just a single grid cell of the environment, but several. This can be done by learning high-level features that are represented as usable functions in the DSL \cite{soemers_spatial_2023}.
In addition, the mutation operator can be improved by training a neural network to predict the probabilities of the production rules in the grammar to guide the mutations into the most promising direction.

%
\bibliography{zotero.bib}

\end{document}